\documentclass[10pt,twocolumn,letterpaper]{article}

\usepackage{cvpr}              % To produce the CAMERA-READY version
\usepackage{graphicx}
\usepackage{amsmath}
\usepackage{amssymb}
\usepackage{booktabs}

\usepackage[pagebackref,breaklinks,colorlinks]{hyperref}

\usepackage[capitalize]{cleveref}
\crefname{section}{Sec.}{Secs.}
\Crefname{section}{Section}{Sections}
\Crefname{table}{Table}{Tables}
\crefname{table}{Tab.}{Tabs.}

\def\cvprPaperID{3} % *** Enter the CVPR Paper ID here
\def\confName{CVPR}
\def\confYear{2023}

\begin{document}

\title{Perceive, Excavate and Purify: A Novel Object Mining Framework for Instance Segmentation}

\author{Jinming Su, Ruihong Yin, Xingyue Chen and Junfeng Luo \\
Meituan \\
{\tt\small \{sujinming, yinruihong, chenxingyue02, luojunfeng\}@meituan.com}
}

\maketitle

\begin{abstract}
Recently, instance segmentation has made great progress with the rapid development of deep neural networks. However, there still exist two main challenges including discovering indistinguishable objects and modeling the relationship between instances. To deal with these difficulties, we propose a novel object mining framework for instance segmentation. In this framework, we first introduce the semantics perceiving subnetwork to capture pixels that may belong to an obvious instance from the bottom up. Then, we propose an object excavating mechanism to discover indistinguishable objects. In the mechanism, preliminary perceived semantics are regarded as original instances with classifications and locations, and then indistinguishable objects around these original instances are mined, which ensures that hard objects are fully excavated. Next, an instance purifying strategy is put forward to model the relationship between instances, which pulls the similar instances close and pushes away different instances to keep intra-instance similarity and inter-instance discrimination. In this manner, the same objects are combined as the one instance and different objects are distinguished as independent instances. Extensive experiments on the COCO dataset show that the proposed approach outperforms state-of-the-art methods, which validates the effectiveness of the proposed object mining framework.
\end{abstract}

\begin{figure}[t]
\centering
\includegraphics[width=1\linewidth,height=11cm]{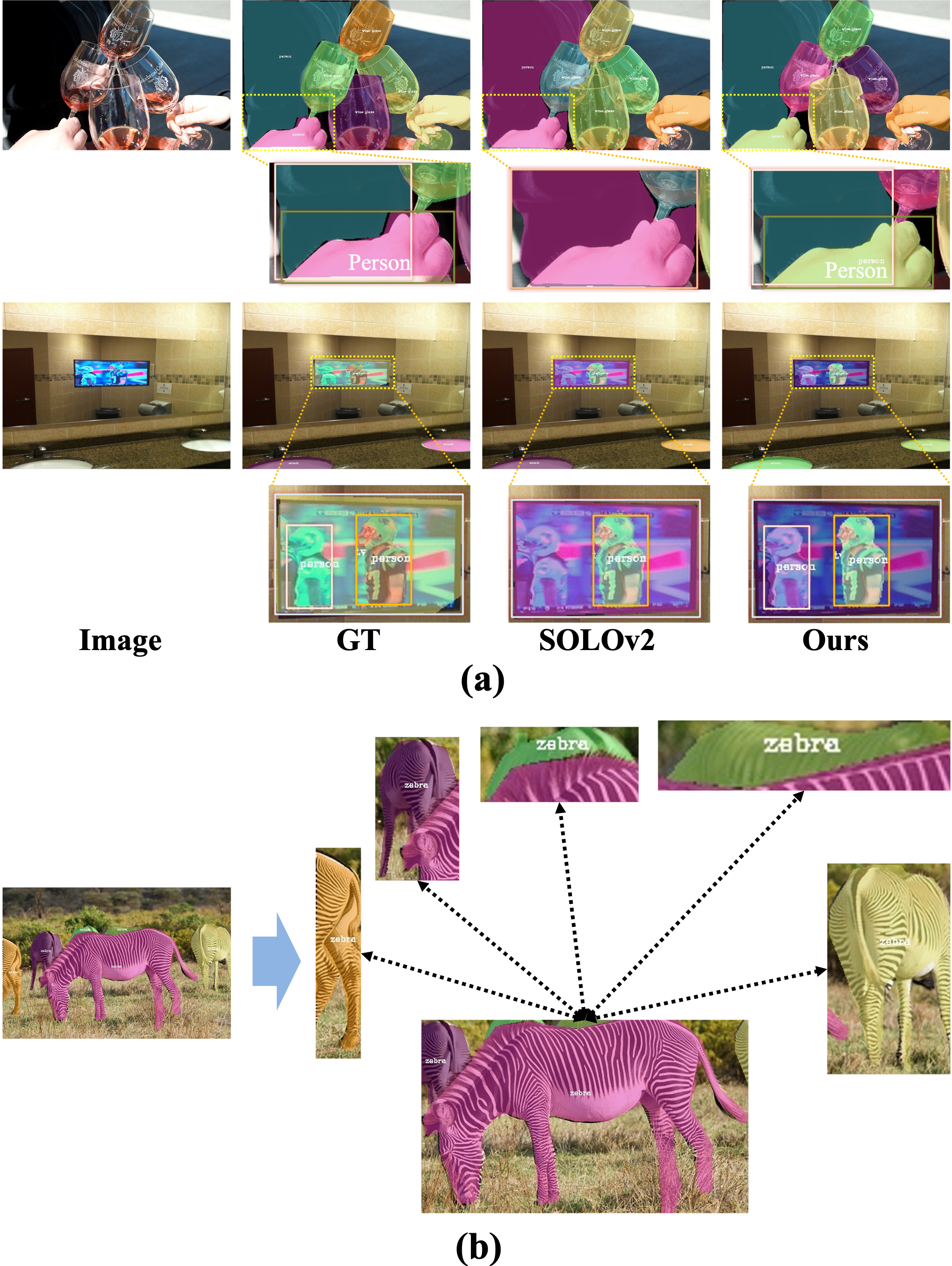}
\caption{Challenges of instance segmentation. (a) Indistinguishable objects. It is difficult for adjacent and overlapping objects (\eg, missed persons from SOLOv2~\cite{wang2020solov2}) to locate and segment.  (b) Underresearched relationship between instances. The relationships (\eg, feature distance)  between different instances are rarely studied while they are important for instance distinguishment.}
\label{fig:motivation}
\end{figure}

\section{Introduction}

Instance segmentation is a fundamental perception task, which aims to detect and segment pre-defined objects at the instance level. Over the past years, the task of instance segmentation has made significant progress with the development of deep neural networks~\cite{krizhevsky2012imagenet} and has a wide range of applications in real-world scenarios such as autonomous driving~\cite{luc2018predicting} and video surveillance~\cite{zhang2020traffic}. 

\begin{figure*}[t]
\centering
\includegraphics[width=1\textwidth,height=7cm]{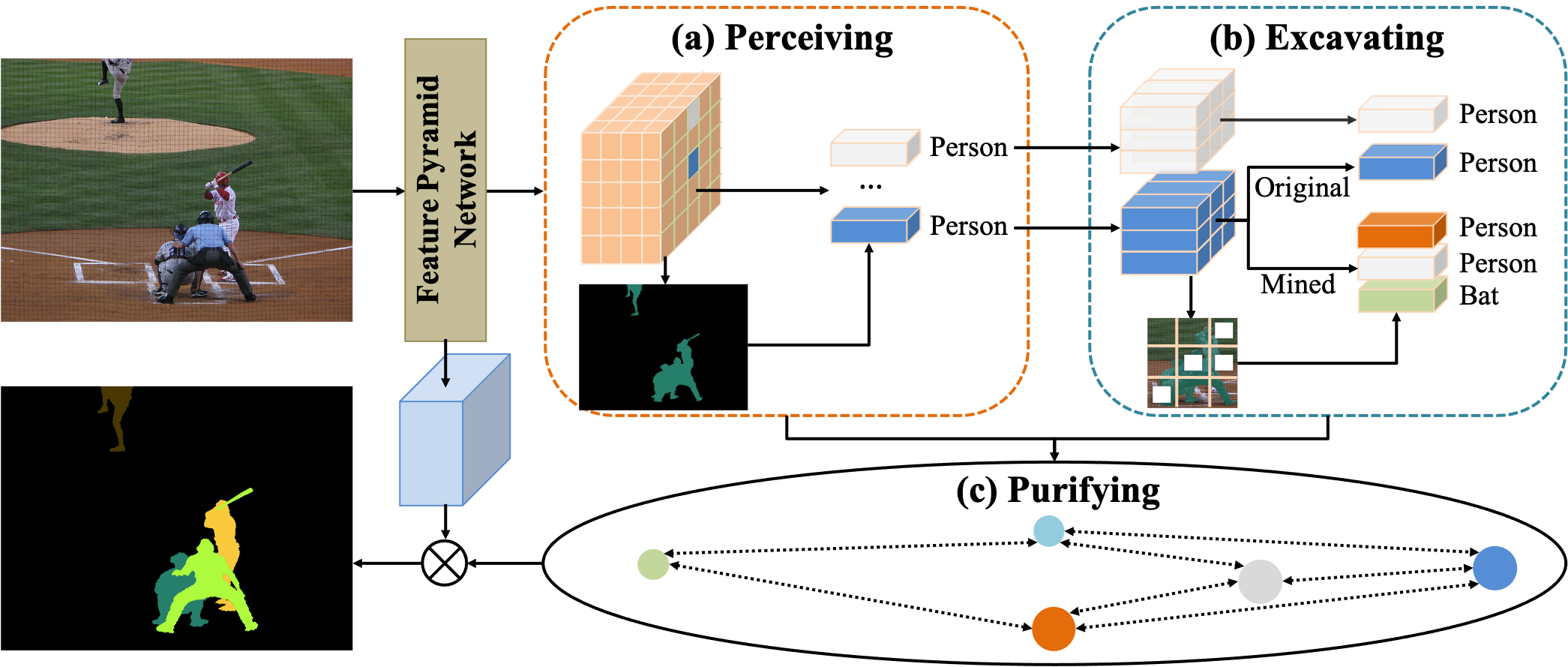}
\caption{Framework of our approach. We first extract pyramid features by the feature pyramid network, which provides features for semantics perceiving and mask learning. Next, a semantics perceiving subnetwork based on semantic segmentation is introduced to capture pixels that may belong to an obvious instance called original instance descriptors with classifications and locations. Then, by learning instances from the region around original descriptors, the object excavating subnetwork is used to discover indistinguishable objects called mined descriptors.  After excavating, all descriptors are modeled in a relationship graph according to feature distances, which can link all instances with each other and obtain final independent instances. Note that $\otimes$ means convolutional operation.}
\label{fig:framework}
\end{figure*}

To address the task of instance segmentation, lots of learning-based methods have been proposed in recent years, achieving impressive performance. For the basic task, the existing methods can be summarized into three categories. (1) Top-down methods~\cite{he2017mask,bolya2019yolact,huang2019mask,tian2019fcos,xie2020polarmask} address the problem based on object detection, \ie, detecting and segmenting the object in the box. (2) Bottom-up methods~\cite{de2017semantic,newell2017associative,gao2019ssap} deal with the problem in a labeling and clustering manner, \ie, learning the per-pixel embeddings first and then clustering them into groups. (3) The latest direct method~\cite{wang2020solo,wang2020solov2} aims at dealing with instance segmentation directly, without dependence on object detection or embedding-and-clustering. However, there still exist two challenges that hinder the development of instance segmentation. First, there usually exists indistinguishable objects in real-world scenarios, especially adjacent and overlapping objects as shown in Fig.~\ref{fig:motivation} (a). Second, The relationships (such as the distance in feature space) between different instances are rarely studied while they are important for instance distinguishment as shown in Fig.~\ref{fig:motivation} (b). Due to these two difficulties, instance segmentation remains a challenging vision task.

To address these two difficulties, many methods have made some efforts. To deal with indistinguishable objects, online hard example mining (OHEM)~\cite{shrivastava2016training,suh2019stochastic} automatically select hard examples to make training more effective and efficient, which boosts the detection performance and is easy to transfer to instance segmentation. In addition, focal loss~\cite{lin2017focal} focuses training on a sparse set of hard examples and prevents the vast number of easy negatives from overwhelming the detector during training. These methods mainly enhance the learning ability of difficult examples at the level of training strategies, which are indirect schemes without awareness at the object level. To address the second difficulty about modeling instance relationship, relation network~\cite{hu2018relation} proposes an object relation module to process a set of objects simultaneously through interaction between their appearance feature and geometry, which allows modeling of their relations. Besides, IRNet~\cite{ahn2019weakly} defines two kinds of relations between a pair of pixels: the displacement between their coordinates and their class equivalence, which are used to train the network effectively. These methods utilize neural networks to implicitly model the instance relationship, but it is still difficult to explore explicit representation for the relationship between instances.

Inspired by these observations and analysis, we propose a novel object mining framework for instance segmentation, as shown in Fig.~\ref{fig:framework}. In this framework, we first introduce semantics perceiving to capture all pixels potentially belonging to an instance, and we select pixels with high category confidence as the original instance descriptors with semantics-level classification and location. In this manner, independent objects are easy to be perceived, and indistinguishable objects mainly exist in the case caused by object gathering. In order to discover indistinguishable objects, we propose an object excavating mechanism. In this mechanism, original descriptors are fed into an instance learning subnetwork to determine whether each pixel is the center of the instance around original descriptors. In this way, indistinguishable objects around origin instances are detected with classification and locations, which are named mined descriptors. After that, we put all the instance descriptors (including both original instance descriptors and mined descriptors) into the instance purifying graph to model the relationship between instances. In this graph, we constrain instance descriptors of the same objects to have the highest similarity but instance descriptors of different objects have little similarity, which can pull the similar instances close and push away different instances. Therefore,  the instance purifying graph can keep intra-instance similarity and inter-instance discrimination. Finally, all independent instances after purifying are utilized to generate final masks by mask learning. Experimental results on the COCO~\cite{lin2014microsoft} dataset achieve state-of-the-art performance, which verifies the effectiveness of the proposed method.

The main contributions of this paper are as follows: 1) we propose an object mining framework for instance segmentation, which achieves state-of-the-art performance and validates the effectiveness of this perceiving, excavating, and purifying pipeline. 2) we put forward the object excavating mechanism, in which indistinguishable objects around original instances are discovered so that hard instances in an image are fully mined. 3) we introduce the instance purifying mechanism to model the relationship between instances, which maintains the intra-instance similarity and inter-instance discrimination.

\section{Related Work}
In this section, we review related works that aim to resolve the challenges of instance segmentation in three aspects.

\subsection{Instance Segmentation}
Instance segmentation is a basic vision task, which requires instance-level and pixel-level predictions. With the development of deep neural networks, many methods~\cite{de2017semantic,newell2017associative,gao2019ssap,wang2020solo,wang2020solov2} 
%he2017mask,bolya2019yolact,huang2019mask,tian2019fcos,xie2020polarmask,
have made great progress in the benchmark dataset~\cite{lin2014microsoft}. For the challenging task, the existing methods can be classified into three folds. First, top-down methods are usually based on object detection to segment objects in a bounding box. For example, 
Mask R-CNN~\cite{he2017mask} extends the object detection network Faster R-CNN~\cite{ren2016faster} by adding a branch for predicting an object mask in parallel with the existing branch for bounding box recognition. 
%And Mask scoring R-CNN~\cite{huang2019mask} contains a network to learn the quality of the predicted instance masks, which calibrates the misalignment between mask quality and mask score, and improves instance segmentation performance by prioritizing more accurate mask predictions during evaluation.
In particular, recent anchor-free methods(\eg, FCOS~\cite{tian2019fcos} and PolarMask~\cite{xie2020polarmask}) also achieve good results with fast inference. The second kind of method is bottom-up methods, which are usually learning the feature of each pixel and clustering pixels into groups in feature space. For example, 
%DLF~\cite{de2017semantic} proposes a discriminative loss function, operating at the pixel level,that encourages a convolutional network to produce a representation of the image that can easily be clustered into instances with a simple post-processing step. And 
SSAP~\cite{gao2019ssap} proposes a pixel-pair affinity pyramid, which computes the probability that two pixels belong to the same instance in a hierarchical manner. The last kind of method is proposed in recent years, which aims at addressing the task of instance segmentation in direct ways. For example, a series of methods, like SOLO~\cite{wang2020solo,wang2020solov2}, assign categories to each pixel within an instance according to the instance's location and size, thus nicely converting instance segmentation into a single-shot classification-solvable problem. In the above methods, various strategies have improved the performance of instance segmentation. However, there still exist two main challenges including discovering indistinguishable objects and modeling the relationship between instances. 

\subsection{Hard Example Mining}
Hard example mining is used to speed up the convergence of neural networks and enhance the discriminative power of the learned features. To reduce the computational overhead of identifying hard examples, existing works have been explored in two directions including an exact search within each mini-batch and an approximate search from the whole dataset~\cite{shrivastava2016training,lin2017focal,jin2018unsupervised,suh2019stochastic}. For example, OHEM~\cite{shrivastava2016training} automatically selects hard examples in a mini-batch to make training more effective and more efficient, which is an intuitive algorithm that eliminates several heuristics and hyper-parameters in common use. In addition, UHEM~\cite{jin2018unsupervised} presents how large numbers of hard negatives can be obtained automatically by analyzing the output of a trained detector on video sequences, and describes simple procedures for mining large numbers of such hard negatives (and also hard positives) from unlabeled video data. Moreover, SCHEM~\cite{suh2019stochastic} makes use of class signatures that keep track of feature embedding online with minor additional cost during training and identifies hard negative example candidates using the signatures. As for the above methods, they mainly enhance the learning ability with hard examples by means of training strategies, which are indirect schemes without awareness at the object level.

\subsection{Object Relation}
It is well believed that modeling relations between objects would make contributions to object recognition. Early works make use of object relations as a post-processing step~\cite{divvala2009empirical}, in which the detected objects are re-scored by considering object relationships. After stepping into the era of deep learning, learning-based object relation is widely explored. For example, relation network~\cite{hu2018relation} introduces an object relation module to process a set of objects simultaneously through interaction between their appearance feature and geometry, thus allowing modeling of their relations.
% This suggests that the relation modules have learned information between objects, which can make compensation for missed information during learning on individual objects. 
Additionally, IRNet~\cite{ahn2019weakly} estimates rough areas of individual instances and detects boundaries between different object classes, which ensures to assign instance labels to the seeds and to propagate them within the boundaries. In this way, the entire areas of instances can be estimated accurately and IRNet is trained with inter-pixel relations on the attention maps, thus no extra supervision is required. In these methods, object relation is studied and proved to be effective for object detection or instance segmentation. Nevertheless, it is not clear which information has been learned in these relation modules.

\section{The Proposed Approach}

\begin{figure}
\centering
\includegraphics[width=1.0\linewidth]{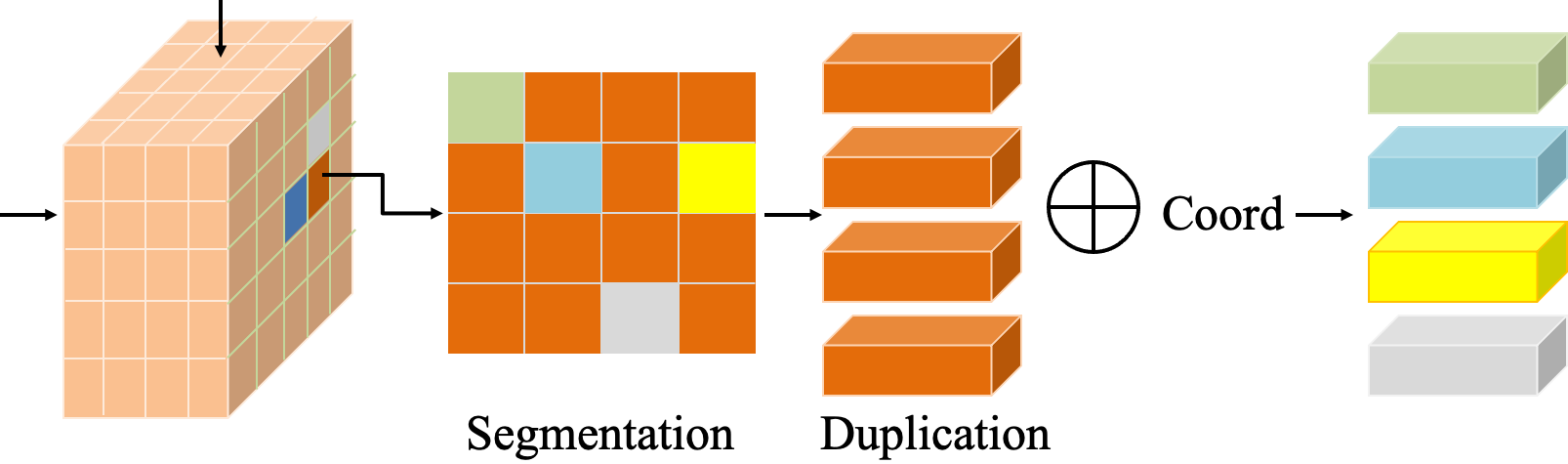}
\caption{Structure of the object excavating subnetwork. The original instance descriptor is fed into the instance-level semantic segmentation to detect objects around original instances. In order to learn instance descriptors of the newly detected instances, instance descriptors of original objects are copied and concatenated with the coordinate to learn new instance descriptors. Note that $\oplus$ means concatenation operation, and Coord means CoordConv.}
\label{fig:mining}
\end{figure}

To address the aforementioned two difficulties (\ie, perceiving indistinguishable objects and modeling the relationship between instances), we propose a novel object mining framework for instance segmentation as depicted in Fig.~\ref{fig:framework}, denoted as \textbf{PEP} (Perceiving, Excavating, and Purifying). In this framework, we first extract common features for further learning. Then, the semantics perceiving subnetwork is introduced to obtain original instance descriptors. The object excavating subnetwork is also proposed to mine indistinguishable objects as mined instance descriptors. Next, the instance purifying mechanism is used to model the relationship between all instances. Finally, all the obtained instance descriptors are fed into mask learning. Details of the proposed approach are described as follows.

\subsection{Feature Extractor}
Inspired by recent methods for panoptic segmentation and instance segmentation~\cite{tian2020conditional,wang2020solo,li2021fully}, we adopt the pyramid convolutional network to extract the common features. As shown in Fig.~\ref{fig:framework}, the proposed method PEP takes feature pyramid network~\cite{he2016deep,lin2017feature} as the feature extractor, which is modified by removing the last two layers (\ie, classification and global average pooling layers) for the pixel-level prediction task. Feature extractor is composed of five stages for feature encoding, named as 
%$\mathcal{C}_{1}(\pi_1), \dots, \mathcal{C}_{5}(\pi_5)$ 
$\mathcal{C}_{s}(\pi_s)$ 
with parameters $\pi_s$, where $s (s=1,2,\dots, 5)$ represent the $s$th stage in the feature extractor.
After the feature extractor, features in each stage are fed into two subnetworks: semantics perceiving subnetworks and mask learning subnetworks. For each subnetwork, there are four $3 \times 3$ convolution layers with 256 kernels to learn the feature transferring. 

\subsection{Semantics Perceiving}

In fact, it is easy to perceive independent objects. Indistinguishable objects mainly exist in the case of object gathering. Therefore, in order to recognize all instances, we first introduce semantics perceiving subnetwork to capture all pixels that may belong to an instance. Specifically, we carry out semantic segmentation to classify each pixel and then choose pixels with high category confidence as the original instance descriptors with semantics-level classification and locations. 

For semantics perceiving subnetwork, we first conduct the common semantic segmentation as semantics perceiving branch $\mathcal{P}_s(\pi_{\mathcal{P}_s})$ with several convolution layers to obtain the feature map with size $C_{\mathcal{P}_s} \times H_s \times W_s$, where $C_{\mathcal{P}_s}, H_s, W_s$ means number of channel, height and width of the feature. In this feature, each pixel $(c,h,w) \in \mathbb{R}^{C_{\mathcal{P}_s} \times H_s \times W_s}$ may represent one object, where $h$ and $w$ are locations of each object in the feature, and $c$ represents the confidence that the object belongs to each class, and the maximum confidence determines the classification of each object. To learn the locations and classification of objects, we expect the output of the semantics perceiving branch to approximate the ground-truth mask (represented as $G_{\mathcal{P}_s}$, which is a one-hot encoding map in the channel dimension and is generated from the ground-truth mask of instance semantics) of classes and locations for each object by minimizing the loss:
\begin{equation}
\mathcal{L}_{\mathcal{P}} = \sum^5_{s=1} CE(\mathcal{P}_s, G_{\mathcal{P}_s}),
\label{eq:preception_loss}
\end{equation}
where $CE(\cdot, \cdot)$ means the cross-entropy loss function with the following formulation:
\begin{equation} \label{eq:binary_cross_entropy}
CE(P, G) = -\sum^{H \times W}_{i} \sum^{C}_{j} G_{i,j}\mathrm{log}P_{i,j},
\end{equation}
where $P_{i,j}$ represents the predicted probabilities that the $i$th ($i \in \mathbb{R}^{H \times W}$) pixel belongs to $j$th ($0 \le j \le C$, 0 represents background) class, $G_{i,j}$ is the ground truth. 

To represent instances better, we adopt descriptors to characterize instances. Toward this end, we introduce instance descriptor extracting branch $\mathcal{D}_s(\pi_{\mathcal{D}_s})$ with several convolution layers. It can learn the instance descriptors represented by the output feature map with size $C_{\mathcal{D}_s} \times H_s \times W_s$, in which these symbols have similar meanings as ones in the object semantics perceiving branch. In the descriptor branch, each object is represented as an instance descriptor $\mathcal{I}$ with size $C_{\mathcal{D}_s} \times 1 \times 1$. In this manner, each object is represented by a $C_{\mathcal{D}_s}$-dimensional vector, which simultaneously ensures the consistency of features and reduces the difficulty of instance representation during learning. By the way, this branch does not directly learn an objective function but is indirectly supervised.

For the proposed semantics and descriptors, object representation is redundant because each non-background pixel can be represented as one object, which will bring extra computation in the next procedures. To reduce the redundancy, only instance descriptors with high category confidence are retained while others are discarded. After that, we get a set of instance descriptors $\mathbb{D}=\{\mathcal{I}_{ind}\}^{N_{ori}}_{ind=1}$ called original instance descriptors, where $\mathcal{I}_{ind}$ means the instance descriptor with index $ind$, and $N_{ori}$ means the number of selected instance descriptors. 

\subsection{Object Excavating}

\begin{figure}
\centering
\includegraphics[width=1.0\linewidth]{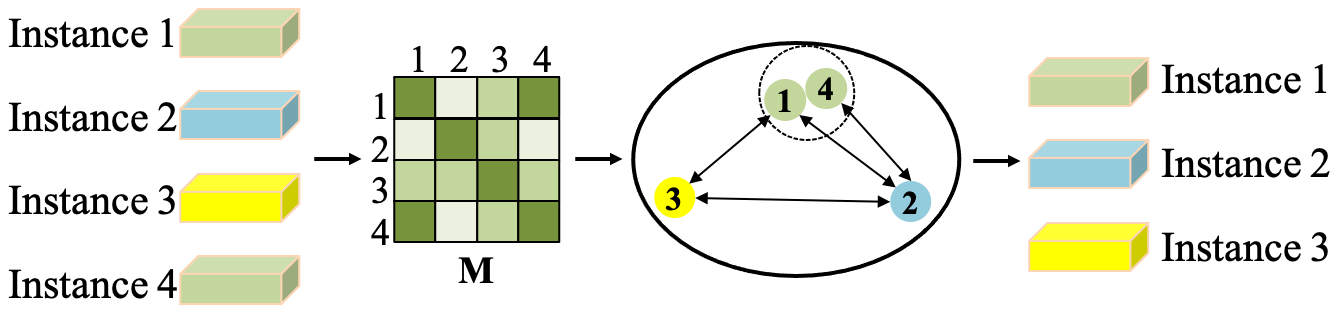}
\caption{Structure of the instance purifying mechanism. Each instance descriptor learns the similarity with other instance descriptors in feature space, ensuring that the similar instances are close and different instances are far away. Then, close instances are merged and different instances are distinguished to get the final instance descriptors.}
\label{fig:linking}
\end{figure}

After semantics perceiving, independent objects can be perceived, and original instance descriptors with classifications and locations are obtained. In fact, original instances can be regarded as easily identifiable objects. Due to object gathering, the rest of the indistinguishable objects mainly exist around the easily identifiable objects. In order to perceive indistinguishable objects, we propose the object excavating subnetwork to mine indistinguishable objects around the original instances.

For mining indistinguishable objects, we introduce the object excavating subnetwork. In the subnetwork, each instance descriptor in $\mathbb{D}$ is fed into an instance-level semantic segmentation subnetwork to detect whether each pixel is the center of a new instance around origin instances as shown in Fig.~\ref{fig:mining}. 
The output ${\mathcal{E}_{ind}}$ of instance-level semantic segmentation around the original instance ${\mathcal{I}_{ind}}$ is easy to learn with several convolutional layers and then supervised by the ground truth of instance-level semantic segmentation $G_{\mathcal{E}_{ind}}$ (one binary map, in which the locations are set to 1 if the pixel is the center of the instance while others representing background are set to 0). To learn the feature map ${\mathcal{E}_{ind}}$, the minimizing optimization objective is as follows:
\begin{equation}
\mathcal{L}_{\mathcal{E}} = \sum^{N_{ori}}_{ind=1} CE({\mathcal{E}_{ind}}, G_{\mathcal{E}_{ind}}).
\label{eq:keypoint_location_loss}
\end{equation}
In this manner, indistinguishable objects around each original instance are detected as the key pixels. 

Then, we learn the corresponding instance descriptor from each key pixel by following operations. Firstly, descriptors of the original instance are copied $N_{\mathcal{E}_{ind}}$ times, which is the same as the number of key pixels. Then, the corresponding coordinate information on each copied feature descriptor is concatenated by CoordConv~\cite{liu2018intriguing}. Then, the new instance descriptors are learned by convolutional layers. At the same time, similar classification learning is carried out for each key pixel ${\mathcal{E}^k_{ind}} (e=1,\dots,N_{\mathcal{E}_{ind}})$ to gain semantic information. To achieve this, we expect the output $\mathcal{P}_{\mathcal{E}^e_{ind}}$ of classification module for the key pixel to approximate its ground truth (represented as $\mathcal{G}_{\mathcal{E}^e_{ind}}$, which is a similar map as $G_{\mathcal{P}_s}$) by minimizing the loss:
\begin{equation}
\mathcal{L}_{\mathcal{P}_{\mathcal{E}}} = \sum^{N_{ori}}_{ind=1} \sum^{N_{\mathcal{E}_{ind}}}_{e=1} CE(\mathcal{P}_{\mathcal{E}^e_{ind}}, \mathcal{G}_{\mathcal{E}^e_{ind}}).
\label{eq:keypoint_classification_loss}
\end{equation}

As a result, through the object excavating subnetwork, we mine objects around each original instance to obtain imperceptible objects with the classification and descriptors. Supposing the number of newly mined objects is $N_{mined}$, we add these new instance descriptors to the original descriptor set and obtain $\mathbb{D}=\{\mathcal{I}_{ind}\}^{N_{all}}_{ind=1}$ where $N_{all} = N_{ori} + N_{mined}$. By the way, we denote the whole object excavating subnetwork as $\mathcal{M}_{s}(\pi_{\mathcal{M}_{s}})$.

\subsection{Instance Purifying}

For the sake of improving the mutual promotion ability between different instances, we introduce the instance purifying mechanism to directly model the relationship between instances as displayed in Fig.~\ref{fig:linking}. In this mechanism, we construct the relationship between each instance and other instances, in which each instance is represented by the instance descriptor $\mathcal{I}_{ind} \in \mathbb{D}$.

To model the relationship, we propose an instance purifying subnetwork to maintain intra-instance similarity and inter-instance discrimination. As shown in Fig.~\ref{fig:linking}, we put all the instance descriptors from the semantic perceiving subnetwork and object excavating subnetwork into the instance purifying graph, in which each node is regarded as an instance descriptor, and the weight of the connection edge between nodes represents the feature distance between two instance descriptors. For the purifying graph, its adjacency matrix $\text{M}$ can represent the relationship between instance descriptors. This matrix has the size of $N_{all} \times N_{all}$. In the matrix, the rows and columns represent the indexes of instances, and each value in the matrix represents the similarities between instances of the corresponding row and column. We drive the matrix $\text{M}$ to approach its ground truth $G_{\text{M}}$ by minimizing the loss function
\begin{equation}
\mathcal{L}_{\text{M}} = CE(\text{M}, G_{\text{M}}).
\label{eq:matrix_loss}
\end{equation}
It is worth noting that modeling pixel-level relationships are usually difficult due to a large amount of computation. However, the number of instance descriptors is usually limited ($N_{all} \le 30$), so the computation is efficient. By graph learning, we constrain the same object to have the highest similarity, but different objects have no similarity. In other words, the graph learning pulls the similar instances close and pushes away different instances, which can keep intra-instance similarity and inter-instance discrimination. 

During inference, we predict the similarity matrix in feature space. Then, we merge the close instances and distinguish different instances to obtain the final instance descriptors. For the convenience of description, we denote this instance purifying subnetwork as $\mathcal{IL}(\pi_{\mathcal{IL}})$.

\begin{table*}[t]
\small
\centering
\setlength{\tabcolsep}{2.5mm}{
\renewcommand\arraystretch{1.0}
\begin{tabular}{c c c c c c c c}
\hline
 & \textbf{Backbone}
 & \textbf{AP}
 & \textbf{AP$_{50}$}
 & \textbf{AP$_{75}$}
 & \textbf{AP$_{S}$}
 & \textbf{AP$_{M}$}
 & \textbf{AP$_{L}$}
 \\
\hline
Two-stage methods \\
\hline
MNC~\cite{dai2016instance}  & ResNet-101 & 24.6 & 44.3 & 24.8 & 4.7 & 25.9 & 43.6 \\
FCIS~\cite{li2017fully}  & ResNet-101 & 29.2 & 49.5 & - & 7.1  & 31.3 & 50.0 \\
Mask R-CNN~\cite{he2017mask}  & ResNet-101 FPN & 35.7 & 58.0 & 37.8 & 15.5 & 38.1 & 52.4 \\
Mask R-CNN~\cite{he2017mask}  & ResNeXt-101 FPN & 37.1 & 60.0 & 39.4 & 16.9 & 39.9 & 53.5 \\
MaskLab+~\cite{chen2018masklab}  & ResNet-101 FPN & 35.4 & 57.4 & 37.4 & 16.9 & 38.3 & 49.2 \\
MS R-CNN~\cite{huang2019mask}  & ResNet-101 FPN & 38.3 & 58.8 & 41.5 & 17.8 & 40.4 & 54.4  \\
HTC~\cite{chen2019hybrid}  & ResNet-101 FPN & 39.7 & 61.8 & 43.1 & 21.0 & 42.2 & 53.5 \\
MS R-CNN$^\dagger$ & ResNet-101 FPN & 39.6 & 60.7 & 43.1 & 18.8 & 41.5 & 56.2 \\
BMask R-CNN~\cite{cheng2020boundary}  & ResNet-101 FPN & 37.7 & 59.3 & 40.6 & 16.8 & 39.9 & 54.6 \\
BMask R-CNN with MS & ResNet-101 FPN & 38.7 & 59.1 & 41.9 & 17.4 & 40.7 & 55.5 \\
BCNet+Faster R-CNN~\cite{ke2021deep} & ResNet-101 FPN & 39.8 & 61.5 & 43.1 & \textbf{\color{red}{\underline{22.7}}} & 42.4 & 51.1 \\
DCT-Mask R-CNN~\cite{shen2021dct} & ResNet-101 FPN & 40.1 & 61.2 & 43.6 & \textbf{\color{red}{\underline{22.7}}}           & 42.7 & 51.8 \\
%RS-Mask R-CNN~\cite{oksuz2021rank} & ResNet-101 FPN & 40.6 & \textbf{\underline{62.8}} & 43.9 & \textbf{\underline{22.8}}           & 43.6 & 52.8 \\
\hline
One-stage methods \\
\hline
YOLACT~\cite{bolya2019yolact} & ResNet-101 FPN & 31.2 & 50.6 & 32.8 & 12.1 & 33.3 & 47.1 \\
TensorMask~\cite{chen2019tensormask} & ResNet-101 FPN & 37.1 & 59.3 & 39.4 & 17.4 & 39.1 & 51.6 \\
ShapeMask~\cite{kuo2019shapemask}  & ResNet-101 FPN & 37.4 & 58.1 & 40.0 & 16.1 & 40.1 & 53.8 \\
PolarMask~\cite{xie2020polarmask} & ResNet-101 FPN & 30.4 & 51.9 & 31.0 & 13.4 & 32.4 & 42.8 \\
CenterMask-W~\cite{wang2020centermask} & Hourglass-104 & 34.5 & 56.1 & 36.3 & 16.3 & 37.4 & 48.4  \\
CenterMask-L~\cite{lee2020centermask} & ResNet-101 FPN & 38.3 & - & - & 17.7 & 40.8 & 54.5 \\
SOLO~\cite{wang2020solo} & ResNet-101 FPN & 37.8 & 59.5 & 40.4 & 16.4 & 40.6 & 54.2\\
BlendMask~\cite{chen2020blendmask} & ResNet-101 FPN & 38.4 & 60.7 & 41.3 & 18.2 & 41.5 & 53.3 \\
MEIns~\cite{zhang2020mask} & ResNet-101 FPN & 33.9 & 56.2 & 35.4 & 19.8 & 36.1 & 42.3 \\
D-SOLO~\cite{wang2020solo} & ResNet-101 FPN & 38.4 & 59.6 & 41.1 & 16.8 & 41.5 & 54.6 \\
SOLOv2~\cite{wang2020solov2} & ResNet-101 FPN & 39.7 & 60.7 & 42.9 & 17.3 & 42.9 & 57.4 \\
CondInst~\cite{tian2020conditional} & ResNet-101 FPN & 39.1 & 60.9 & 42.0 & 21.5 & 41.7 & 50.9 \\
BCNet+FCOS~\cite{ke2021deep} & ResNet-101 FPN & 39.6 & 61.2 & 42.7 & 22.3 & 42.3 & 51.0 \\
BoundaryFormer~\cite{lazarow2022instance}& ResNet-101 FPN & 39.4 & 60.9 & 42.6 & 22.1 & 42.0 & 51.2 \\
%SOTR~\cite{guo2021sotr} & 2021 & ResNet-101 FPN & 40.2 & 61.2 & 43.4 & 10.3 & 59.0 & 73.0 \\
\hline

Ours & ResNet-101 FPN & \textbf{\color{red}{\underline{40.9}}} & \textbf{\color{red}{\underline{61.5}}} & \textbf{\color{red}{\underline{44.3}}} & 18.2 & \textbf{\color{red}{\underline{44.4}}} & \textbf{\color{red}{\underline{59.7}}} \\

%Ours & - & ResNet-101 FPN & \textbf{\underline{39.6}} & \textbf{\underline{60.8}} & \textbf{\underline{42.7}} & 17.0 & \textbf{\underline{43.0}} & \textbf{\underline{57.6}} \\
%Ours$^\dagger$ & - & ResNet-101 FPN & 41.8 & 62.8 & 45.4 & 18.1 & 45.0 & 61.4 \\
\hline
\end{tabular}
}
\caption{Comparisons with state-of-the-art methods on COCO \emph{test-dev2017} dataset. The best result is in \textbf{\color{red}{\underline{bold}}} fonts.}
\label{tab:performance_on_coco}
\end{table*}

\subsection{Mask Learning}

After obtaining the instance descriptors $\mathcal{I}_{ind}$, we use a general feature learning branch $\mathcal{B}_s(\pi_{\mathcal{B}_s})$ to produce a general feature with size $C_{\mathcal{B}_s} \times H_s \times W_s$ to represent the feature space of object mask learning, which provides shared pixel-level feature bases for all instance descriptors. In this way, all descriptors are convolved with the general features $\mathcal{B}$ to obtain the instance segmentation result as follows.
\begin{equation}
M_{ind} = \mathcal{I}_{ind} \otimes \mathcal{B},
\label{eq:promotion_network_loss}
\end{equation}
where $\otimes$ means the convolutional operation. Instance segmentation result $M_{ind}$ is expected to approximate the ground-truth
masks $G_{M_{ind}}$ by minimizing the loss
\begin{equation}
\mathcal{L}_{M} = \sum_{ind=1}^{N_{all}} CE(M_{ind}, G_{M_{ind}}).
\label{eq:mask_loss}
\end{equation}

By taking the losses of Eqs.(\ref{eq:preception_loss}),(\ref{eq:keypoint_location_loss}), (\ref{eq:keypoint_classification_loss}), (\ref{eq:matrix_loss}), and (\ref{eq:mask_loss}), the overall learning objective can be formulated as follows:
\begin{equation}
\begin{aligned}
\min _{\mathbb{P}} \mathcal{L}_\mathcal{P} + \alpha \mathcal{L}_\mathcal{E} + \beta \mathcal{L}_{\mathcal{P}_{\mathcal{E}}} + \gamma \mathcal{L}_\text{M} + \delta \mathcal{L}_M,
\end{aligned}
\label{eq:final_loss}
\end{equation}
where $\mathbb{P}$ is the set of $\{\pi_s, \pi_{\mathcal{B}_s}, \pi_{\mathcal{P}_s}, \pi_{\mathcal{D}_s}, \pi_{\mathcal{M}_s}\}^5_{s=1}$ and $\pi_\mathcal{IL}$ for convenience of presentation.
%where $\mathbb{P}$ is the set of $\{\{\pi_s, \pi_{\mathcal{P}_s}, \pi_{\mathcal{D}_s}, \pi_{\mathcal{B}_s}, \pi_{\mathcal{M}_s}\}^5_{s=1}$ and $\pi_\mathcal{IL}$ for convenience of presentation.

\section{Experiments and Results}

\begin{figure*}[t]
\centering
\includegraphics[width=1\linewidth,height=14.5cm]{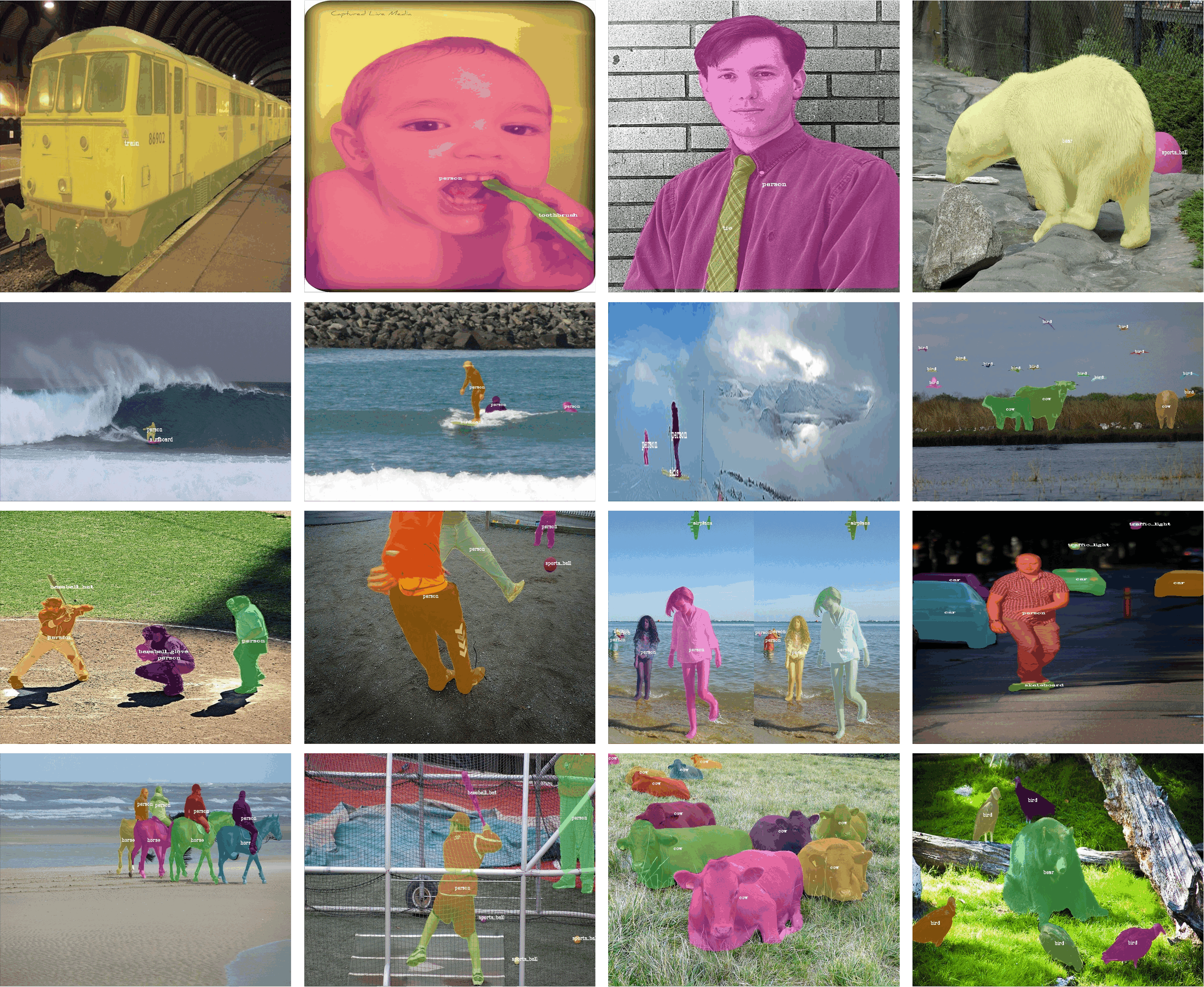}
\caption{Visualization examples of the proposed approach.}
\label{fig:result}
\end{figure*}

\begin{table*}[t]
\small
\centering
\setlength{\tabcolsep}{1.1mm}{
\renewcommand\arraystretch{1.0}
\begin{tabular}{c c c c c c c }
\hline
 & \textbf{AP}
 & \textbf{AP$_{50}$}
 & \textbf{AP$_{75}$}
 & \textbf{AP$_{S}$}
 & \textbf{AP$_{M}$}
 & \textbf{AP$_{L}$}
 \\
\hline
Baseline &               38.5 & 59.1 & 41.4 & 17.2 & 42.8 & 57.4 \\
Baseline + excavating &  39.8 & 60.9 & 42.1 & 17.5  & 43.3 & 58.7 \\
Baseline + purifying &  40.1 & 60.8 & 42.3 & 17.6 & 43.9 & 58.4 \\
\hline
PEP & \textbf{\color{red}{\underline{40.6}}} & \textbf{\color{red}{\underline{61.0}}} & \textbf{\color{red}{\underline{43.2}}} & \textbf{\color{red}{\underline{17.7}}} & \textbf{\color{red}{\underline{44.3}}} & \textbf{\color{red}{\underline{59.8}}} \\
\hline
\end{tabular}
}
\caption{Performance of different settings of the proposed method on COCO \emph{val2017} dataset.}
\label{tab:ablation_on_coco}
\end{table*}

\subsection{Experimental Setup} 
\paragraph{Dataset.} To evaluate the performance of the proposed method, we carry out all experiments on COCO dataset~\cite{lin2014microsoft}. 
%COCO consists of complex everyday scenes and contains common objects in their natural context, which is split into training, validation, and test (note the annotation of the testing set is withheld for official benchmarking purposes) sets. 
As set in previous work~\cite{lin2017feature,he2017mask}, we train our network using the union of 118k train images \emph{train2017}, conduct ablation studies on the 5k validation images \emph{val2017}, and report final results on \emph{test-dev2017} for comparison with state-of-the-art methods.

\paragraph{Metrics.} We report the results with the standard COCO metrics including AP (averaged over IoU thresholds), AP$_{50}$, AP$_{75}$, and AP$_{S}$, AP$_{M}$ and AP$_{L}$  (AP at different scales). 
%Unless otherwise noted, AP is evaluating using mask IoU.

\paragraph{Training and Inference.} Our proposed method can be trained end-to-end. We apply the stochastic gradient descent algorithm with a weight decay of 0.0001 and momentum of 0.9 to optimize the loss in Eq.~\ref{eq:final_loss}. In the optimization process, parameters in the feature extractor are initialized by the pre-trained ResNet-101~\cite{he2016deep} model, whose output dimensions for the second to fifth stages are 256, 512, 1024, and 2048 respectively. COCO \emph{train2017} dataset is used as the training set. The shorter side of the input image is resized to 800 pixels, and the longer side is less than or equal to 1333 pixels. We train our network for 36 epochs with a mini-batch of 16 images (8 GPUs $\times$ 2 mini-batch). “Step” learning rate policy is employed for all experiments. The initial learning rate is 0.01, which is decreased by a factor of 10 at 27 and 33 epoch respectively. In addition, we empirically set the weight $\alpha=\beta=\gamma=\delta=1$ in Eq.~\ref{eq:final_loss}. The dimension $C_{\mathcal{D}_s}$ of instance descriptors is 256. To match the dimension in instance descriptor, the dimension $C_{\mathcal{B}_s}$ of the general feature $B$ is 256, too. During inference, all supervisions are removed, and the final result is obtained from the output of mask learning.

\subsection{Comparisons with State-of-the-art Methods}
We compare our approach with about 26 state-of-the-art algorithms in instance segmentation, including two-stage methods (MNC~\cite{dai2016instance}, FCIS~\cite{li2017fully}, Mask R-CNN~\cite{he2017mask}, MaskLab+~\cite{chen2018masklab}, MS R-CNN~\cite{huang2019mask}, HTC~\cite{chen2019hybrid}, DCT-Mask R-CNN~\cite{shen2021dct}, BMask R-CNN~\cite{cheng2020boundary}, and BCNet+Faster R-CNN~\cite{ke2021deep}), and one-stage methods ( YOLACT~\cite{bolya2019yolact}, TensorMask~\cite{chen2019tensormask}, ShapeMask~\cite{kuo2019shapemask}, PolarMask~\cite{xie2020polarmask}, CenterMask-W~\cite{wang2020centermask}, CenterMask-L~\cite{lee2020centermask}, SOLO~\cite{wang2020solo}, SOLOv2~\cite{wang2020solov2}, CondInst~\cite{tian2020conditional} and  BCNet+FCOS~\cite{ke2021deep}) 
%SOTR~\cite{guo2021sotr}
on COCO \emph{test-dev2017} dataset.

% MEIns~\cite{zhang2020mask}, BlendMask~\cite{chen2020blendmask}

To prove the effectiveness of our method, we report our results based on ResNet-101 with FPN, as listed in Tab.~\ref{tab:performance_on_coco}. Comparing the proposed method with state-of-the-art methods, we can see that our method consistently outperforms both one-stage and two-stage methods across almost all metrics. For AP, compared with the second-best methods DCT-Mask R-CNN in two-stage methods and BCNet+FCOS in one-stage methods, our method is noticeably improved by 0.8\% (from 40.1\% to 40.9\%) and 1.3\% (from 39.6\% to 40.9\%) respectively. Additionally, it is worth noting that compared with DCT-Mask R-CNN, our method is significantly better on AP$_{M}$ (+1.7\%), AP$_{L}$ (+7.9\%) respectively. As for AP$_{S}$ in one-stage methods, we are below the methods (BCNet+Faster R-CNN and DCT-Mask R-CNN), which conduct more computations for fine feature learning and decoding. However, our method has fewer computational overheads. Compared with BCNet+FCOS, our method gains an 8.7\% higher result in AP$_{L}$. Overall, these observations present the effectiveness and robustness of our proposed method and validate that the object mining framework is useful for the task of instance segmentation. 

% DCN+FPN~\cite{dai2017deformable}

Some examples generated by our approach are shown in Fig.~\ref{fig:result}. We can see that instances can be distinguished and segmented with accurate location and precise shape by the proposed method in both easy (like large objects) and complex situations(small objects, occluded and densely distributed scenes). These visualizations indicate that the proposed method has a good ability of object excavating and instance purifying. Besides, it also shows the effectiveness and superiority of the proposed object mining method.

\subsection{Ablation Analysis}
To validate the effectiveness of different components, we carry out several experiments on COCO \emph{val2017} dataset and compare the performance variations of our framework PEP.

Firstly, to investigate the effectiveness of the proposed object excavating mechanism, we conduct ablation experiments and introduce two different models for comparisons. The first setting only consists of the feature extractor, semantics perceiving, and the masking learning, which is regarded as the ``Baseline'' model. In addition, we carry out another model (``Baseline+excavating'') by adding object excavating subnetwork to Baseline. 
The comparisons of above two models are listed in Tab.~\ref{tab:ablation_on_coco}. As for the metric AP, the object excavating mechanism can improve the performance of Baseline from 38.5\% to 39.8\% (+1.3\%). We also observe that the object excavating mechanism can consistently improve the performance of Baseline model on all metrics by 0.3\%-1.8\%, which verifies the effectiveness of the proposed object excavating mechanism.

Secondly, to explore the effectiveness of the instance purifying mechanism, we conduct another experiment by combining the instance purifying subnetwork with Baseline as “Baseline+purifying”. 
From the third row of Tab.~\ref{tab:ablation_on_coco}, we find that the performance of instance segmentation can be consistently improved (from 38.5\% to 40.1\% on AP), which validates that the instance purifying mechanism is effective.  

Thirdly, comparing the proposed framework PEP with other models in Tab.~\ref{tab:ablation_on_coco}, we can also find that if combining both object excavating and instance purifying mechanisms in PEP, our method can gain further performance. This shows that our proposed object excavating and instance purifying mechanisms are compatible with each other.

\section{Conclusion}
In this paper, we rethink the difficulties that hinder the development of instance segmentation and propose an object mining framework. In this framework, we introduce a novel object excavating mechanism to mine the objects around each instance, which ensures to fully excavate hard samples. In addition, the instance purifying mechanism is introduced to model the relationship between instances, which pulls similar instances close and pushes away different instances. Extensive experiments on benchmark datasets validate the effectiveness of the proposed approach and show that the perspective of object mining is useful for the task.

%%%%%%%%% REFERENCES
{\small
\bibliographystyle{ieee_fullname}
\bibliography{RefOMLNet}
}

\end{document}